\documentclass[journal]{IEEEtran}

\usepackage{amsmath, lipsum}
\usepackage[justification=centering]{caption}
\usepackage{graphicx}
\usepackage{amsfonts,amssymb}
\usepackage{cuted}
\usepackage{algorithm}
\usepackage{algpseudocode}
\usepackage{booktabs}
\usepackage{cite}
\usepackage[numbers,sort&compress]{natbib}
\usepackage{url}
\usepackage{color}

\begin{document}

\title{Exploring Driving Behavior for Autonomous Vehicles Based on Gramian Angular Field Vision Transformer}

\author{Junwei You,
        Ying Chen*,
        Zhuoyu Jiang,
        Zhangchi Liu,
        Zilin Huang,
        Yifeng Ding,
        Bin Ran
\thanks{*Corresponding author: Ying Chen (y-chen@northwestern.edu)

This work is accepted by IEEE Transactions on Intelligent Transportation Systems.}}

\markboth{IEEE TRANSACTIONS ON INTELLIGENT TRANSPORTATION SYSTEMS}%
{Shell \MakeLowercase{\textit{et al.}}: Bare Demo of IEEEtran.cls for IEEE Journals}

\maketitle

\begin{abstract}
Effective classification of autonomous vehicle (AV) driving behavior emerges as a critical area for diagnosing AV operation faults, enhancing autonomous driving algorithms, and reducing accident rates. This paper presents the Gramian Angular Field Vision Transformer (GAF-ViT) model, specifically designed for analyzing AV driving behavior. The GAF-ViT model is developed upon a novel integration of three key components: GAF Transformation Module, which transforms multivariate driving behavior representative sequences into multi-channel images; Channel Attention Module, which prioritizes relevant behavioral features to enhance classification effectiveness; and Multi-Channel ViT Module, which employs advanced image recognition techniques to accurately classify the resulting multi-channel driving behavior images. This framework not only facilitates detailed analysis of complex multivariate driving behavioral data but also leverages the capabilities of vision-based pattern recognition methods to uncover subtle driving behavior nuances. Experimental evaluation on the Waymo Open Dataset of trajectories demonstrates that the proposed model outperforms baseline models, achieving state-of-the-art performance. Furthermore, an ablation study effectively validates the efficacy of individual modules within the model.
\end{abstract}


\section{Introduction}
\IEEEPARstart{O}{ver} the last decade, the automotive industry has increasingly focused on advancing autonomous vehicle (AV) technology to enhance road safety and substantially mitigate accidents, as distractions and errors from drivers are attributed to an estimated 94\% of all incidents \cite{muzahid2023multiple}. Concurrently, research suggests that the behavior of AVs should mimic that of human drivers to ensure coherent understanding by drivers of other vehicles and adherence to human cognitive patterns \cite{wu2022human}. In light of this, numerous scholars have devoted themselves to developing models that minimize the behavioral gap between AVs and human drivers \cite{zhang2018human, fu2019human, lu2019personalized, chen2022towards, huang2024human}. Nonetheless, a study conducted by the Insurance Institute for Highway Safety (IIHS), which analyzed over 5,000 accidents reported by law enforcement through a nationwide survey, reveals that even with a human-like driving approach and sensors offering a 360-degree view of the surroundings, AVs may only circumvent one-third of accidents. This includes those caused by detection failures and incapacitation, while the majority -- encompassing those arising from speeding, intervention from other drivers, etc.,\-- remain inevitable \cite{mueller2020humanlike}. The study suggests that if AVs exhibit the same level of aggressiveness as human drivers on the road, crashes will continue to occur. Alternatively, setting autonomous driving systems to adopt a conservative approach in mixed traffic flow may reduce the likelihood of fatal crashes but has the potential to generate bottlenecks. Such a prudent approach could frustrate or irritate other drivers and significantly elevate the probability of rear-end collisions, especially in complex decision-making scenarios like intersections and four-way stops \cite{mit_news}.

Analogous to traditional vehicles, where effectively identifying human drivers' behavior serves as an additional informational reference for surrounding vehicles -- enabling proactive decision-making and reducing the crash probabilities -- \cite{hong2020driver} identifying and classifying behavior for AVs is crucial. This not only has the potential to guide the evaluation of the stochasticity and stability of autonomous driving algorithms but also fosters the improvement of functionality.

However, to the best of the authors' knowledge, the majority of research to date has principally focused on the classification of traditional drivers' behavior, often neglecting the varied behaviors exhibited by AVs \cite{shahverdy2020driver, melnyk2023driver, zhang2022study, kuroki2021semi, zhou2019analysis, hong2019rules}. Another segment of the research places a greater emphasis on the motion of AVs in a specific spatiotemporal context - either predicting the vehicle's state in the ensuring planning horizon based on historical data or making optimal decisions in response to environmental changes \cite{miglani2019deep, tang2022prediction, wang2021intelligent, mandal2020motion, luo2020probabilistic, jeong2020surround}. In light of these insights, this paper primarily focuses on the relatively stable and comprehensive behavioral characteristics displayed by autonomous vehicles in mixed traffic flow. 

Specifically, a Gramian Angular Field Vision Transformer (GAF-ViT) model is proposed for analyzing AVs' driving behavior. The Gramian Angular Field (GAF) is a mathematical representation of time series data, capturing the pairwise angular relationships between data points within a time series. This approach decodes time series data into images, thereby facilitating the application of computer vision methods for time series classification \cite{wang2015imaging}. GAF encompasses two types: Gramian Angular Summation Field (GASF) and Gramian Angular Difference Field (GADF). In this study, the driving behavior of AVs is represented by a multivariate feature matrix, which is constructed from different feature sequences. Each feature sequence is transformed into two images in terms of GASf and GADF. The final imagery, representative of the behavioral multivariate matrix, includes the concatenation of two images for each feature, followed by the concatenation of all image sets corresponding to all features. The constructed multi-channel images are then classified through the Vision Transformer (ViT), a powerful deep learning architecture that extends the Transformer model from natural language processing (NLP) to computer vision tasks \cite{dosovitskiy2020image}. In addition, a channel attention structure is incorporated and applied to the generated multi-channel images to differentiate the importance of distinct feature images. The Channel attention mechanism was originally developed to enhance the importance of certain channels or feature maps while suppressing others within a neural network layer \cite{hu2018squeeze}. Through the effective classification of AV driving behaviors, the model assists developers in recognizing various AV driving behaviors and identifying the hazardous ones of certain vehicles, thereby contributing to the timely adjustments or updates of autonomous driving algorithms or control of AVs to prevent accidents. The proposed GAF-ViT model is evaluated on the processed Waymo Open Dataset of trajectories and achieves the best performance among benchmark models. The major contributions of this study include: 

1) An approach based on GAF is proposed to visualize driving behavior features for AVs, effectively converting intricate driving behavior data into a visually interpretable format.

2) An innovative GAF-ViT model is introduced, capable of transforming multivariate feature sequences representing behavior into multi-channel images. This enables efficient classification of driving behavior for AVs through effective recognition of these images.

3) Domain knowledge is seamlessly integrated into the model through a channel attention mechanism, which highlights the most relevant driving behavior features. This integration significantly augments the model’s performance.

The remainder of this paper is organized as follows. Section \uppercase\expandafter{\romannumeral2} reviews existing methods for classifying traditional human drivers' behavior. Section \uppercase\expandafter{\romannumeral3} presents the overall structure as well as each specific module of the developed model in detail. Numerical experiments and results are discussed in Section \uppercase\expandafter{\romannumeral4}. Section \uppercase\expandafter{\romannumeral5} concludes the paper and addresses possible future works.

\section{Literature Review}
One major approach to classifying and analyzing drivers' behavior is based on the modality of data, which underscores the variety of data sources pertinent to studying driving behavior. The following section provides a breakdown of methodologies tailored to each type of data.

\subsection{Vision-Based Methods} 
These methods primarily use visual data from cameras mounted inside or outside the vehicle. In-vehicle cameras primarily capture driver-related information, including facial expressions, gaze, and head posture, thereby becoming crucial for monitoring driver behavior, detecting distractions, and identifying driver fatigue \cite{xing2017identification, zhang2020driver, liu2020leveraging}. Conversely, the integration of exterior cameras, which concentrate on the vehicle's surroundings to detect lane departures and monitor the traffic environment, contributes to classifying driving behaviors. These behaviors relate to lane-keeping, adherence to traffic rules, collision avoidance, and interactions with the surrounding environment, providing a more holistic view of driver actions and prevailing road conditions \cite{fan2022hybrid, fridman2019advanced}. However, it is noteworthy that, although camera-based methods exhibit exemplary performance, especially in intricate traffic scenarios, they may be cost-prohibitive and computationally demanding due to the financial implications of hardware installation and maintenance, as well as the computational requirements for processing visual camera data \cite{wang2017much}. 

\subsection{Trajectory-Based Methods}
\subsubsection{General Trajectory-Based Analysis}
General trajectory methods for driving behavior analysis leverage data collected from various inertial or positioning sensors to construct vehicle trajectories, including speed, acceleration, gyroscopic and accelerometric measurements, GPS, etc. This data type is important in classifying driving behaviors such as abrupt maneuvers, aggressive driving, and adherence to speed limits. Models typically constructed to utilize this data for classifying drivers' behaviors encompass Kalman filter-based classifiers \cite{wu2016novel}, machine learning-based neural networks such as Support Vector Machine (SVM) and Random Forest (RF) \cite{lopez2012driver, zhang2016driver, xie2019maneuver, malik2023framework}, Long-Short Term Memory (LSTM)-based deep learning architectures \cite{saleh2017driving, moukafih2019aggressive, lee2023privacy, gong2023sifdrivenet}, and Transformer-based models \cite{sharma2023kernelized, vyas2022transdbc}. Typically, trajectory data is structured as a multivariate time series, incorporating various driving behavior features for optimal utility. They are generally highly accurate, real-time, and weather-resistant, affording a comprehensive understanding of driving behaviors \cite{saleh2017driving}. In this study, trajectory data is exclusively used for model training and evaluation.

\subsubsection{Car-Following Behaviors}
Car-following behavior is a critical aspect of trajectory-based methods for driving behavior analysis, focusing on the dynamics between a leading vehicle and a following vehicle. Recent advancements have utilized machine learning and deep learning techniques to enhance the accuracy and robustness of car-following behavior predictions. For instance, Qin et al. developed a CNN-LSTM model that combines convolutional neural networks (CNNs) with LSTM networks to analyze and predict car-following behavior using trajectory data, demonstrating superior accuracy and generalization ability compared to classical models \cite{qin2023cnn}. Another study by Fan et al. employed an LSTM to investigate the impact of driving memory on car-following behavior, highlighting the significance of historical driving data in predicting future behaviors \cite{fan2019car}. Additionally, Qin et al. proposed a car-following model based on a combination of LSTM and Transformer networks, with the focus on reconstructing input features from trajectory data to improve model performance under data loss scenarios \cite{qin2023research}. Furthermore, Li et al. explored the identification of automated vehicles using car-following trajectory data and developed learning-based models that significantly enhance the accuracy of vehicle classification in mixed traffic environments \cite{li2023automated}. These advanced models capture the temporal dependencies and complex interactions inherent in car-following scenarios from trajectory data, contributing to a more effective and adaptive traffic management system. By incorporating car-following behavior when studying driving behavior, researchers can better address the nuances of driving behavior, ultimately enhancing the accuracy of behavior classification and prediction in automated driving systems.

\subsection{Smartphone-Based Methods}
Smartphone-based methods for classifying driving behaviors utilize sensors such as accelerometers and GPS equipped in smartphones to capture pertinent data. These methods offer notable advantages, including cost-effectiveness, portability, and real-time monitoring, attributable to the widespread use of smartphones \cite{eren2012estimating, carlos2019smartphone, lindow2019driver, khodairy2021driving, brahim2022machine}. Nonetheless, they may encounter challenges related to data accuracy, sensor variability across different smartphone models, privacy concerns, limitations due to available sensor types, and considerations related to battery life \cite{mantouka2021smartphone}. Overall, while smartphone-based approaches prove to be well-suited for large-scale monitoring and individual driver supervision, they may necessitate additional considerations for tasks that require high precision and specialized analysis.

\section{Methodology}
\subsection{Structure of GAF-ViT Model}
The proposed GAF-ViT model can process any generic multivariate sequential data indicative of specific autonomous driving behavior and generate the corresponding class label. The input multivariate data, denoted as ${F}$, can be defined as follows: 
\begin{equation}
    F = [F_1, F_2, ..., F_i, ..., F_n]
\end{equation}
where ${F_i}$ represents a series of driving behavior features such as speed, acceleration, headway, etc., to effectively identify the specific behavior that all features, collectively referred to as $F$, the GAF-ViT model incorporates three essential modules: GAF Transformation Module, Channel Attention Module, and Multi-Channel ViT Module, as illustrated in Figure 1. 
\begin{figure}[htp]
    \centering
    \includegraphics[width=4.5cm]{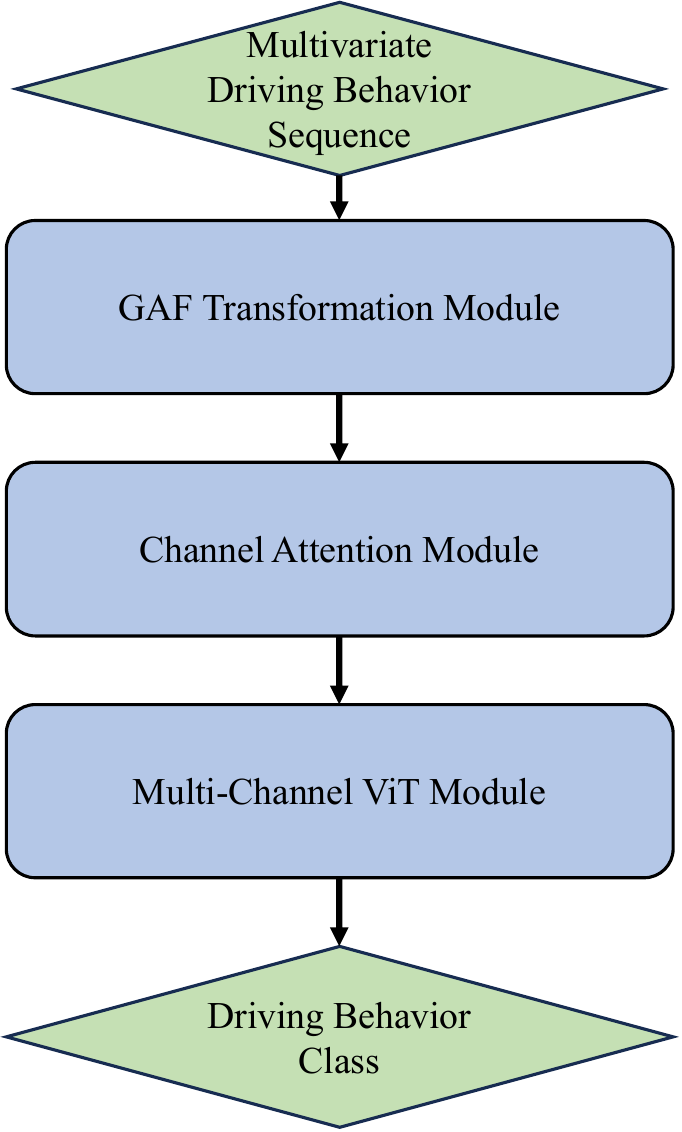}
    \caption{Overview of GAF-ViT Model}
    \label{fig:fig1}
\end{figure}
Specifically, the GAF Transformation Module converts the input $F$ into a multi-channel image. The Channel Attention Module is employed to enhance or attenuate the significance of specific feature channels and is connected with the Multi-Channel ViT Module, which ultimately classifies the image into a specific driving behavior. Subsequent sections will provide detailed illustrations of each module.

\subsection{GAF Transformation Module}
The Gramian Angular Field (GAF) technique is predominately utilized in time series analysis and signal processing \cite{wang2015imaging}. It articulates time series data in a way that captures the intrinsic temporal relationships and patterns. Previous studies indicate that GAF is particularly adept at analyzing and visualizing the cyclic or periodic behavior inherent in time series data \cite{xu2020human, hong2023monitoring}. Moreover, the GAF transformation uniquely facilitates the application of general vision-based pattern recognition models to time series analysis. This leverages their spatial pattern detection capabilities to uncover intricate temporal relationships and features within the data. Thus, it offers a novel approach that amalgamates time-domain analysis with visual pattern recognition techniques \cite{garibo2023gramian}. As noted in the previous section, there are two types of GAF: GASF and GADF. Generally, given a univariate time series data denoted as vector ${F_i}=[f_{i1}, f_{i2}, ..., f_{ij}, ..., f_{im}]$, the normalized vector $\tilde{F_i}$ of ${F_i}$ can be determined as follows:
\begin{equation}
    \tilde{F_i} = \frac{f_{ij}-\min F_i}{\max F_i - \min F_i}
\end{equation}
where the element $\tilde f_{ij}$ in $\tilde{F_i}$ is within $[0, 1]$. The scaled vector $\tilde{F_i}$ can then be expressed in polar coordinates, transitioning from Cartesian coordinates, as follows: 
\begin{equation}
    \phi_{ij} = \arccos \tilde f_{ij}, r = \frac{j}{m}
\end{equation}
where $\phi_{ij}$ represents the angular value and $r$ denotes the radius in the polar coordinate. Finally, two types of GAF can be calculated according to the following equations:
\begin{small}
\begin{equation} 
\begin{split}
   &GASF_i = \cos (\phi_{ij}+\phi_{ik})\\
           &= \left[                
                \begin{array}{cccc}   
                \cos (\phi_{i1}+\phi_{i1}) & \cos (\phi_{i1}+\phi_{i2}) & \cdots & \cos (\phi_{i1}+\phi_{im})\\
                \cos (\phi_{i2}+\phi_{i1}) & \cos (\phi_{i2}+\phi_{i2}) & \cdots & \cos (\phi_{i2}+\phi_{im})\\
                \vdots & \vdots & \ddots & \vdots\\
                \cos (\phi_{im}+\phi_{i1}) & \cos (\phi_{im}+\phi_{i2}) & \cdots & \cos (\phi_{im}+\phi_{im})\\
                \end{array}
            \right]\\
          &= \cos \phi_{ij}\cos \phi_{ik} - \sqrt {1-\cos^2 \phi_{ij}}\sqrt {1-\cos^2 \phi_{ik}}\\
          &= \tilde f_{ij}\tilde f_{ik} - \sqrt{1-\tilde f_{ij}^2}\sqrt{1-\tilde f_{ik}^2}\\
          &= \tilde{F_i}^\top\tilde{F_i} - \sqrt{\mathbf{I}-\tilde{F_i}^2}^\top\sqrt{\mathbf{I}-\tilde{F_i}^2}      
\end{split}
\end{equation}
\begin{equation} 
\begin{split}
   &GADF_i = \sin (\phi_{ij}-\phi_{ik})\\
          &= \left[               
                \begin{array}{cccc}   
                \sin (\phi_{i1}-\phi_{i1}) & \sin (\phi_{i1}-\phi_{i2}) & \cdots & \sin (\phi_{i1}-\phi_{im})\\
                \sin (\phi_{i2}-\phi_{i1}) & \sin (\phi_{i2}-\phi_{i2}) & \cdots & \sin (\phi_{i2}-\phi_{im})\\
                \vdots & \vdots & \ddots & \vdots\\
                \sin (\phi_{im}-\phi_{i1}) & \sin (\phi_{im}-\phi_{i2}) & \cdots & \sin (\phi_{im}-\phi_{im})\\
                \end{array}
            \right]\\
          &= \cos \phi_{ik}\sqrt {1-\cos^2 \phi_{ij}} - cos \phi_{ij}\sqrt {1-\cos^2 \phi_{ik}}\\
          &= \tilde f_{ik}\sqrt{1-\tilde f_{ij}^2} - \tilde f_{ij}\sqrt{1-\tilde f_{ik}^2}\\
          &= \sqrt{\mathbf{I}-\tilde{F_i}^2}^\top\tilde{F_i} - \tilde{F_i}^\top\sqrt{\mathbf{I}-\tilde{F_i}^2}
\end{split}
\end{equation}
\end{small}
Hence, for each feature vector of $F$, two images are obtained in terms of GASF and GADF. These two images are then stacked together. The necessity of stacking GASF and GADF together, rather than using them separately, arises from the fact that the transformation of either GASF or GADF is not injective. This means that different time series can produce identical GASF or GADF images due to the symmetry of cosine and sine functions. For instance, reversing the sign of every point in a time series results in a new time series with each value negated. However, this reversal does not alter the resulting GASF image, as the cosine of an angle plus its supplementary angle yields the same value. Hence, employing both sine and cosine functions to generate two types of GAF images and then combining them ensures that the angular values are uniquely determined, thereby addressing the non-injectivity issue. Ultimately, all the stacked images for all feature vectors are concatenated, forming the multi-channel image. The whole process of the GAF Transformation Module is illustrated in Figure 2.  
\begin{figure*}
    \centering
    \includegraphics[width=17cm]{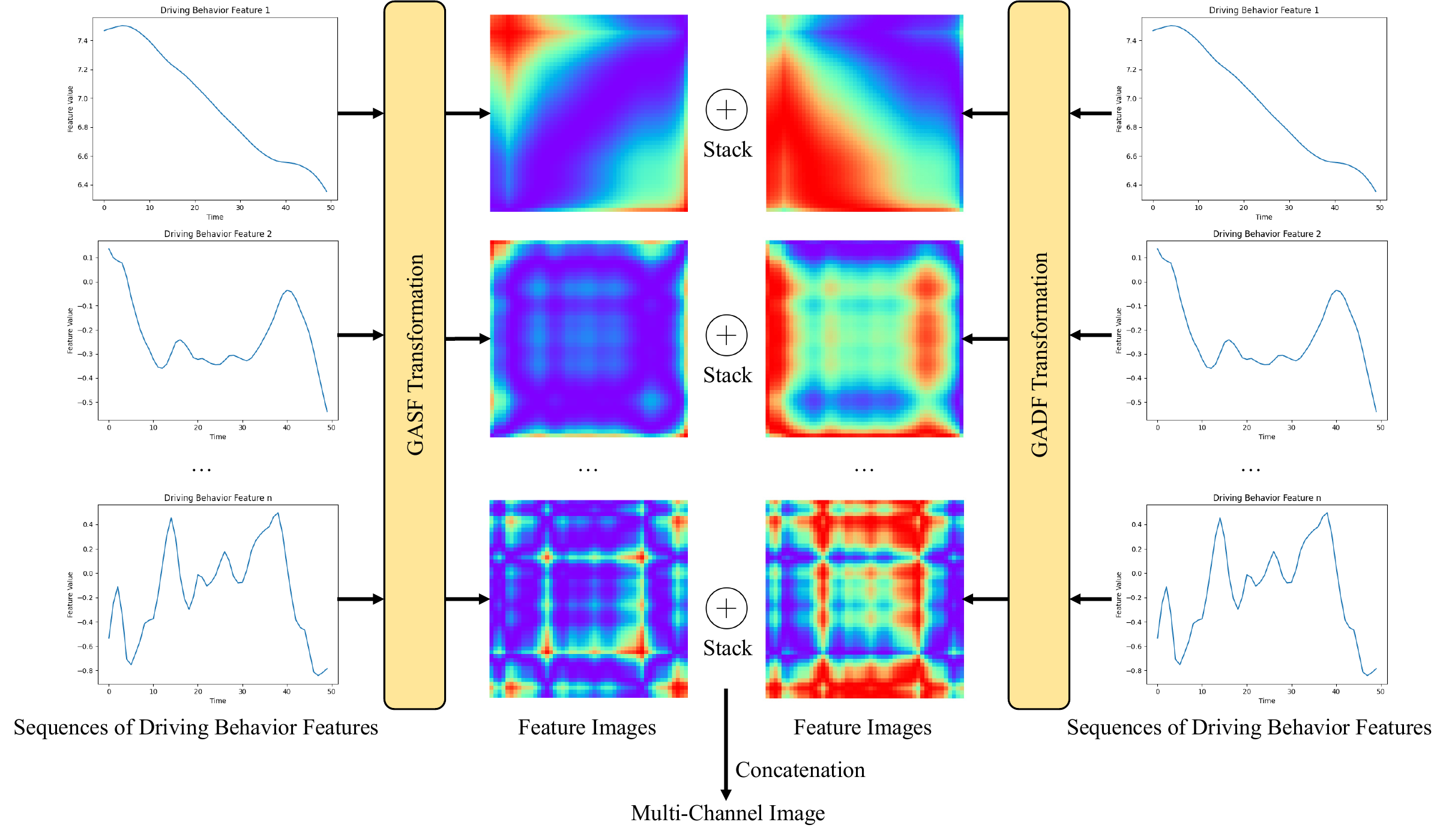}
    \caption{Workflow of GAF Transformation Module}
    \label{fig:fig2}
\end{figure*}

\subsection{Channel Attention Module}
Before training a classification model to understand driving behavior from generated multi-channel images, the channel Attention Module (CAM) is employed to facilitate the learning of distinct weights for different channels along the channel dimension while maintaining uniform weights across spatial dimensions \cite{hu2018squeeze, jin2022delving}. Intuitively, various behavior features uniquely contribute to defining vehicle driving behavior. For instance, speed may be a more direct indication of a car's threatening state compared to acceleration. As images transformed from feature sequences are allocated to their respective channels in the multi-channel image, the application of varying learnable weights to distinct channels quantifies this contribution, thereby improving classification accuracy. Figure 3 illustrates the structure of the Channel Attention Module.
\begin{figure}
    \centering
    \includegraphics[width=9cm]{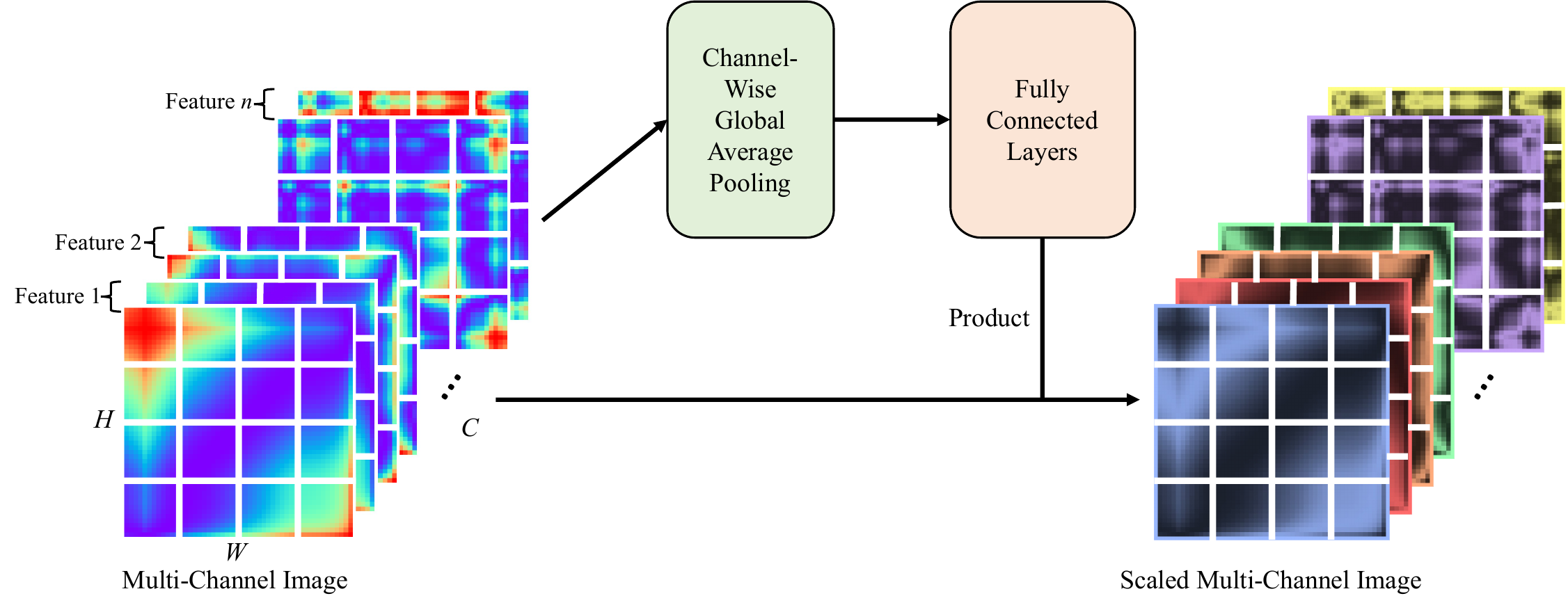}
    \caption{Structure of the Channel Attention Module}
    \label{fig:fig3}
\end{figure}
Specifically, the approach encompasses three primary steps. Let the constructed multi-channel image be denoted as tensor $V$, characterized by width $W$, height $H$, and depth $C$. Firstly, global average pooling is applied across both width and height dimensions to compress $V$, thereby encoding the comprehensive spatial feature of each channel into a singular feature. This operation results in an intermediate output tensor $u$ of dimensions $1\times 1\times C$. The squeezing process is mathematically formulated as follows: 
\begin{equation}
    u = f_{squeeze}(V) = \frac{1}{H\times W}\sum\limits_{i=1}^{H}\sum\limits_{j=1}^{W} V(i, j)
\end{equation}
Sequentially, two fully connected (FC) layers are used to discern the dependencies across channels, with weights given by the FC layers being normalized via a Sigmoid function, denoted as $\sigma$. The function confines the weights within the $[0, 1]$ range while ensuring their sum equals 1. The process to generate the final weights $U$ for the channels can be formulated as follows:
\begin{equation}
    U = \sigma (W_2\alpha (W_1u))
\end{equation}
where $W_1$ and $W_2$ represent weights generated within the FC layers and $\alpha$ denotes the ReLU activation function. 

Finally, by multiplying the final weights by their corresponding channels, the weights are effectively integrated into the input multi-channel image, leading to a newly scaled multi-channel image that will serve as the input of the classification model. 

\subsection{Multi-Channel ViT Module}
In recent years, Transformer models, particularly those based on self-attention mechanisms, have emerged as the foremost choice for Natural Language Processing (NLP) tasks \cite{vaswani2017attention, han2021transformer, rao2021msa, kitaev2020reformer} due to their exceptional performance. The Vision Transformer (ViT) applies Transformer architecture to image processing \cite{dosovitskiy2020image}, involving the segmentation of an image into numerous patches and subsequently processing these linearly arranged patch sequences as input to the Transformer model. In this context, the concept of image patches is similar to the concept of tokens in NLP tasks. This study illustrates the structure of the adapted Multi-Channel ViT Module, utilized for classifying multi-channel images representative of driving behavior and scaled as delineated in the previous section, in Figure 4. 
\begin{figure*}[htp]
    \centering
    \includegraphics[width=18cm]{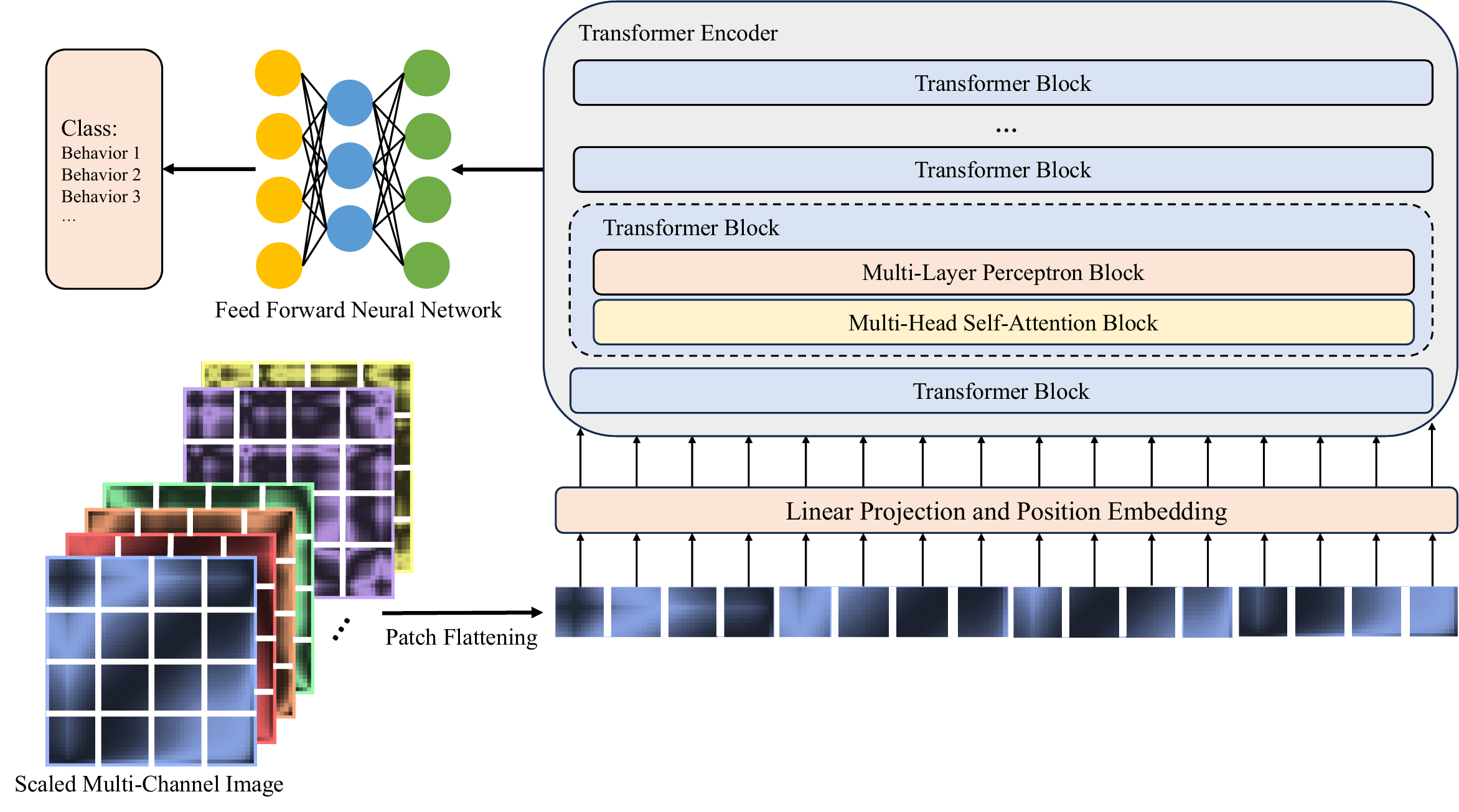}
    \caption{Structure of Multi-Channel ViT Module}
    \label{fig:fig4}
\end{figure*}
Given an input image, denoted as $X\in \mathbb{R}^{H\times W\times C}$, it is reshaped into a sequence of flattened patches, expressed as $X_p\in \mathbb{R}^{N\times (P^2C)}$, where $H$, $W$ and $C$ represent the height, width and number of channels of the scaled multi-channel image, respectively, $P$ refers to the patch size, while $N = \frac{HW}{P^2}$ indicates the number of patches. Subsequently, each patch is linearly embedded into a lower-dimensional feature space, with an extra class embedding and position embedding appended to the patch embedding. The embedding processes can be expressed as follows:
\begin{equation}
    z_0 = [X_{class}; X_p^1E; X_p^2E; ...; X_p^iE; ...; X_p^NE] + E_{position}
\end{equation}
where $X_{class}$ denotes the embedded class label, $X_p^i$ represents the $i$-th patch, $E$ defines the linear patch embedding function, with $E_{position}$ representing the position embedding function, and $z_0$ being the output of the embedding operations. The Transformer encoder comprises a sequential array of Transformer blocks, and the total number of Transformer blocks is $L$. Each Transformer block encompasses a multi-head self-attention (MSA) block, followed by a multi-layer perceptron (MLP) block, as shown in the following equations:
\begin{equation}
    z_l^{'} = MSA(LN(z_{l-1})) + z_{l-1}, l = 1, ..., L
\end{equation}
\begin{equation}
    z_l = MLP(LN(z_l^{'})) + z_l^{'}, l = 1, ..., L
\end{equation}
where $LN$ denotes the layer normalization function, $z_l$ and $z_l^{'}$ correspond to the outputs from the MLP and MSA blocks, respectively. Residual connections are also incorporated subsequent to each MSA and MLP block. The MLP block comprises two fully connected layers, utilizing a GeLU activation function. Ultimately, the predicted score for each driving behavior class is generated through a feed-forward neural network connected to the Transformer encoder, as expressed in the following equation:
\begin{equation}
    y = LN(z_L)
\end{equation}
\subsection{GAF-ViT Model Operation}
Based on the elaboration of each module presented in previous sections, this section demonstrates a general implementation of the proposed GAF-ViT model. It accepts the multivariate driving behavior feature matrix $F$ as input and yields the predicted behavior class label $\hat{Y}$ as output.
\begin{algorithm}
\captionsetup{}
\caption{Algorithm of GAF-ViT} 
\textbf{Input:} Multivariate driving behavior feature matrix $F=[F_1, F_2, ..., F_i, ..., F_n]$
\begin{algorithmic}[1]
\For{$i=1, ..., n$}
\State $\tilde{F_i} \gets \text{Normalize $F_i$}$
\State $\text{GASF}(\tilde{F_i}) = \tilde{F_i}^\top\tilde{F_i} - \sqrt{\mathbf{I} - \tilde{F_i}^2}^\top \sqrt{\mathbf{I} - \tilde{F_i}^2}$
\State $\text{GASF}(\tilde{F_i}) = \sqrt{\mathbf{I}-\tilde{F_i}^2}^\top\tilde{F_i} - \tilde{F_i}^\top\sqrt{\mathbf{I}-\tilde{F_i}^2}$
\State $\text{$GAF_i$} \gets \text{Concatenate [\text{GASF}($\tilde{F_i}$), \text{GADF}($\tilde{F_i}$)]}$
\EndFor
\State $\text{Multi-Channel image $V$} \gets  \text{Concatenate all $GAF_i$}$
\State $\text{Attention Weights $U$} \gets \text{\textbf{Channel Attention} ($V$)}$
\For{each channel \(C_j\) in \(V\) \textbf{and} each weight \(u_j\) in \(U\)}
\State Scaled channel \(C_j^{'}\) = \(u_j\times C_j\)
\EndFor
\State Scaled multi-channel image \(X\) = [\(C_1^{'}\), \(C_2^{'}\), ..., \(C_j^{'}\), ..., \(C_{2n}^{'}\)]
\State $\text{Class label $\hat{Y}$} \gets \text{\textbf{ViT}($X$)}$
\end{algorithmic}
\textbf{Output:} Predicted driving behavior class label $\hat{Y}$ 
\end{algorithm}

\section{Experiment}
\subsection{Dataset}
The proposed GAf-ViT model is trained and evaluated on a trajectory dataset derived from the Waymo Open Dataset \cite{hu2022processing}. This dataset has undergone preprocessing steps, including outlier removal and denoising. The refined dataset contains three key features of AV driving behaviors: speed, acceleration, and jerk. All three features are used in this study for both model training and testing purposes. The dataset includes 2,704 trips, with most trips lasting approximately 20 seconds in duration and recorded at a time interval of 0.1 seconds. By filtering out trips with a consistent speed of 0 or less, 2,695 meaningful trajectories are retained for analysis. Figure 5 visualizes position ($m$), speed ($m/s$), acceleration ($m/s^2$), and jerk ($m/s^3$) variations over time ($s$) within a trajectory from the processed dataset.
\begin{figure}[htp]
    \centering
    \includegraphics[width=8.5cm]{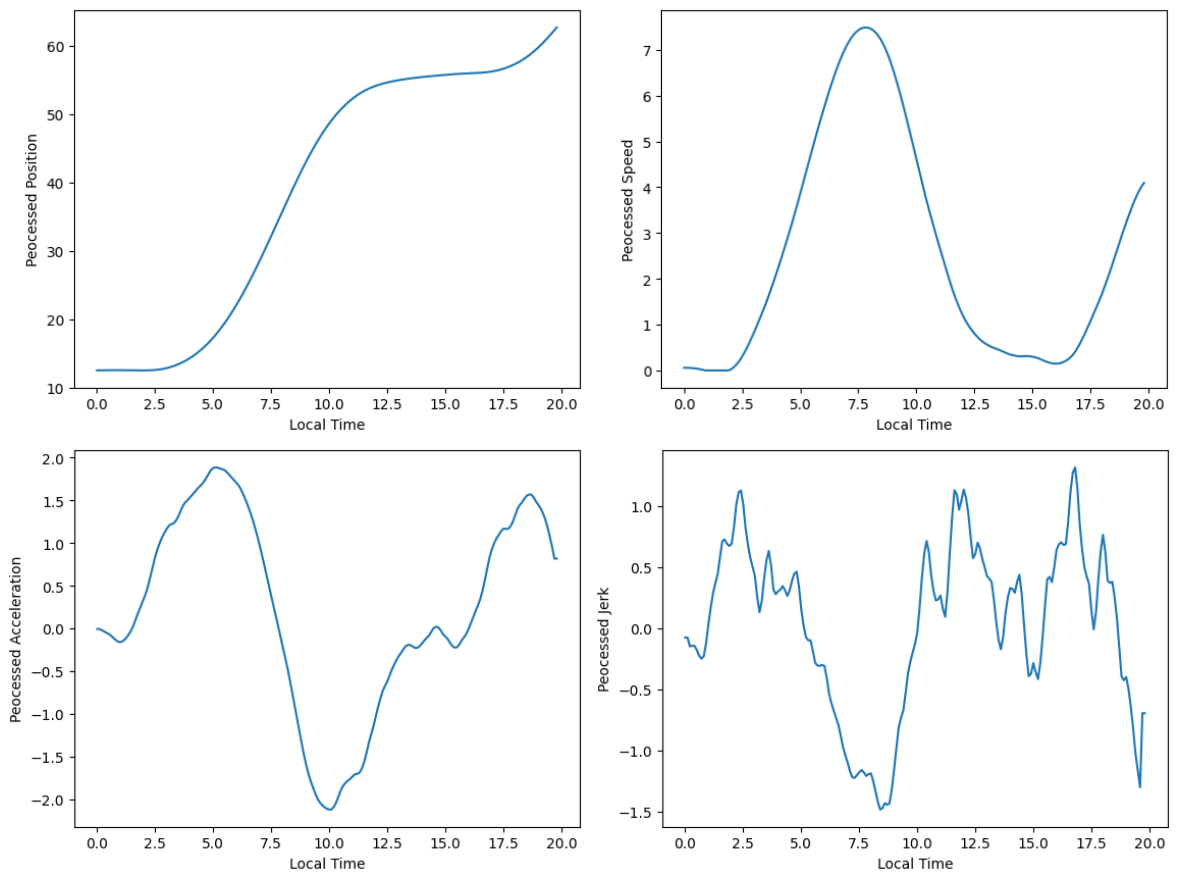}
    \caption{Visualization of a Trajectory}
    \label{fig:fig4}
\end{figure}
In addition, it's important to note that the lengths of trajectories in the dataset vary. The most prevalent trip lengths constitute approximately 86.76\% of the 2,695 trajectories and are 198 or 199 data points long. Consequently, for the experimental dataset, trajectories with a length of 198 or 199 were selected. These sorted trajectories were then split into two trips, and the final data point of those with a length of 199 was removed from each, resulting in each trip having a length of 99. This process yielded 4,674 trajectories as the final dataset prepared for input into the developed model in this study.

\subsection{Implementation Details}
The proposed GAF-ViT model is developed using PyTorch. Some implementation details are described as follows: 

1) \textit{Data Preprocessing:} Each trajectory concatenates three corresponding feature sequences: speed, acceleration, and jerk, into a 3-dimensional matrix, and there are 4,674 matrices in total. To discern the behavior classes represented by the constructed feature matrices, the QuickBundles (QB) algorithm is applied to cluster these multivariate matrices \cite{garyfallidis2012quickbundles}, while the elbow method aids in determining the optimal number of clusters. The QB algorithm is particularly well-suited for multivariate streamline clustering due to its efficiency and adaptability to high-dimensional data \cite{blazquez2021clustering}. Unlike many traditional clustering methods that may struggle with the complexity and computational demands of processing multivariate time series data like trajectories, QB efficiently handles large datasets by grouping streamlines based on a fast approximation of their similarity \cite{kumyaito2020trajectory}. Thus, the hierarchical nature of QB allows for a scalable approach to clustering, enabling the handling of diverse driving behaviors and patterns with varying levels of detail. The detailed implementation of QB-based multivariate driving behavior clustering is outlined in Algorithm 2. Specifically, in this method, the distance between the last row vectors of two matrices, known as endpoint features, is computed for simplicity. This simplification allows for a faster computation of similarity or dissimilarity between streamlines, focusing on their overall orientation rather than their detailed shape or path. The cosine distance computation then evaluates how parallel or divergent these vectors are, serving as a proxy for the similarity between the matrices' overall directions. The cosine distance for two given endpoint feature vectors can be calculated following the equations below:
\begin{equation}
    \cos(\alpha) = \min(\max(\cos(\frac{v_i \cdot v_{cj}}{\|v_i\| \cdot \|v_{cj}\|}), -1), 1) 
\end{equation}
\begin{equation}
    d(v_i, v_{cj}) = \frac{\arccos(\cos(\alpha))}{\pi}
\end{equation}
where $v_i$ and $v_{cj}$ are the endpoint feature vectors of two matrices.
\begin{algorithm}
\captionsetup{}
\caption{Algorithm of QB-based Multivariate Driving Behavior Clustering}
\textbf{Input:} Set of multivariate matrices $M = \{m_1, m_2, ..., m_n\}$, optimal threshold $\theta$
\begin{algorithmic}[1] 
\State Initialize clusters $C$ with the first matrix as a singleton cluster
\For{each matrix $m_i \in M$}
    \State Extract the endpoint feature vector $v_i$ from $m_i$
    \For{each cluster centroid $c_j \in C$}
        \State Extract the endpoint feature vector $v_{cj}$ from $c_j$
        \State Compute cosine distance $d(v_i, v_{cj})$
    \EndFor
    \If{minimum $d(v_i, v_{cj}) \leq \theta$}
        \State Assign $m_i$ to the closest cluster $C_j$
        \State Update centroid of $C_j$ considering $m_i$
    \Else
        \State Create a new cluster with $m_i$ as the initial member
    \EndIf
\EndFor
\State \textbf{Note:} The optimal threshold $\theta$ is pre-determined by the elbow method.
\end{algorithmic}
\textbf{Output:} Clusters $C$
\end{algorithm}

QB clustering effectively reveals four types of driving behaviors across all trajectory samples. Table \uppercase\expandafter{\romannumeral1} presents the number of samples and basic statistics of speed mean ($m/s$), acceleration standard deviation ($m/s^2$), and jerk standard deviation ($m/s^3$) for each class. It is notable that the statistical values in Table \uppercase\expandafter{\romannumeral1} provide a class indication; actual speed, acceleration, and jerk values within a trajectory can fluctuate significantly, both within and across classes. Therefore, based on each class's indicators, interpretations are also provided in Table \uppercase\expandafter{\romannumeral1}. With the identified driving behavior class labels, training of the proposed GAF-ViT model can enable efficient recognition of specific driving behavior classes through the multivariate input feature matrix. 
\begin{table*}[!htbp]
\centering
\caption{Statistics of Driving Behavior Classes}\label{tab:aStrangeTable}
\begin{tabular}{cccccc}
\toprule
Class&Number of Samples&Speed Mean&Acceleration Standard Deviation &Jerk Standard Deviation&Interpretation\\
\bottomrule
0& 3052& 7.05& 0.96& 0.43&Aggressive\\
1& 715& 6.66& 0.87& 0.39&Assertive\\
2& 430& 2.91& 0.80& 0.19&Conservative\\
3& 477& 4.25& 0.75& 0.28&Moderate\\
\toprule
\end{tabular}
\end{table*}

2) \textit{GAF Transformation Module:} Feeding the 3D multivariate driving behavior feature matrices into the GAF Transformation Module, corresponding multi-channel images are generated. For visualization purposes, Figure 6 shows two examples of transformed images, utilizing both GASF and GADF methods, for each feature of an input matrix across each class. It can be seen that even among different samples from the same class, the imaging of identical features, such as speed, demonstrates a consistent pattern.   
\begin{figure*}
    \centering
    \includegraphics[width=18cm]{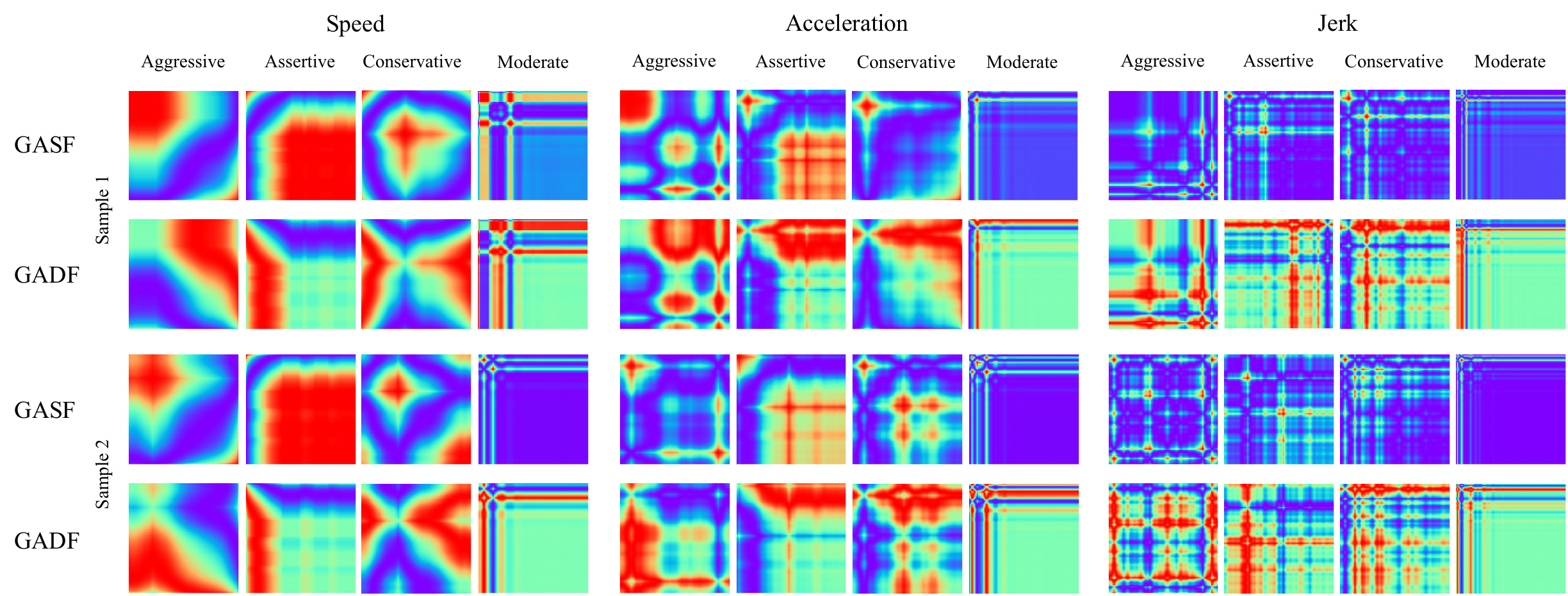}
    \caption{Comparison of Imaging of Driving Behavior Features Across Different Driving Behavior Classes}
    \label{fig:fig6}
\end{figure*}

3) \textit{Channel Attention Module:} In this study, it is anticipated that the scaled multi-channel images generated from the Channel Attention Module maintain the same size as the input multi-channel images. Consequently, the reduction ratio is set to 1 to preserve the consistent dimensionality of the channel-wise information from input to output.

4) \textit{Multi-Channel ViT Module:} As addressed above, each trajectory used in this study has a length of 99, and three features are considered. Thus, each input matrix measures $99\times 3$, yielding a transformed multi-channel image size of $99\times 99$. Accordingly, the selected hyperparameters for the Multi-Channel ViT Module are shown in Table \uppercase\expandafter{\romannumeral2}.
\begin{table}[!htbp]
\centering
\caption{Multi-Channel ViT Module Hyperparameters}\label{tab:aStrangeTable}
\begin{tabular}{cc}
\toprule
Hyperparameter&Value\\
\bottomrule
Number of Patches&11\\
Number of Classes&4\\
Last Dimension of Output Tensor After Linear Transformation&128\\
Number of Transformer Blocks&4\\
Number of Heads in the MSA Layer&4\\
Dimension of the MLP layer&128\\
\toprule
\end{tabular}
\end{table}

5) \textit{Training Setting:} { In this study,  all trajectories are randomly split into 80\% for training, 10\% for validation, and 10\% for testing.} A batch size of 8 and an initial learning rate 1e-5 are employed. The model is trained for 50 epochs utilizing the Cross-Entropy loss function and the AdamW optimizer\cite{loshchilov2019decoupled}, with weight decay regularization applied to prevent over-fitting. 

6) \textit{Evaluation Metrics:} In effectively assess the model performance, the metrics of Accuracy, Precision, Recall, and the F1 score are used, each of which can be calculated using the following equations:
\begin{equation}
    Accuracy = \frac{TP+TN}{TP+TN+FP+FN}
\end{equation}

\begin{equation}
    Precision = \frac{TP}{TP+FP}
\end{equation}

\begin{equation}
    Recall = \frac{TP}{TP+FN}
\end{equation}

\begin{equation}
    F1 = \frac{2\times Precision\times Recall}{Precision+Recall}
\end{equation}
where TP, TN, FP and FN refer to the number of true positive, true negative, false positive, and false negative cases, respectively. Figure 7 illustrates the curves representing variations in loss and accuracy over epochs for both training and validation sets. Notably, the performance begins to converge after approximately 50 epochs on the training set, while on the validation set, it converges after roughly 15 epochs. 
\begin{figure}
    \centering
    \includegraphics[width=9cm]{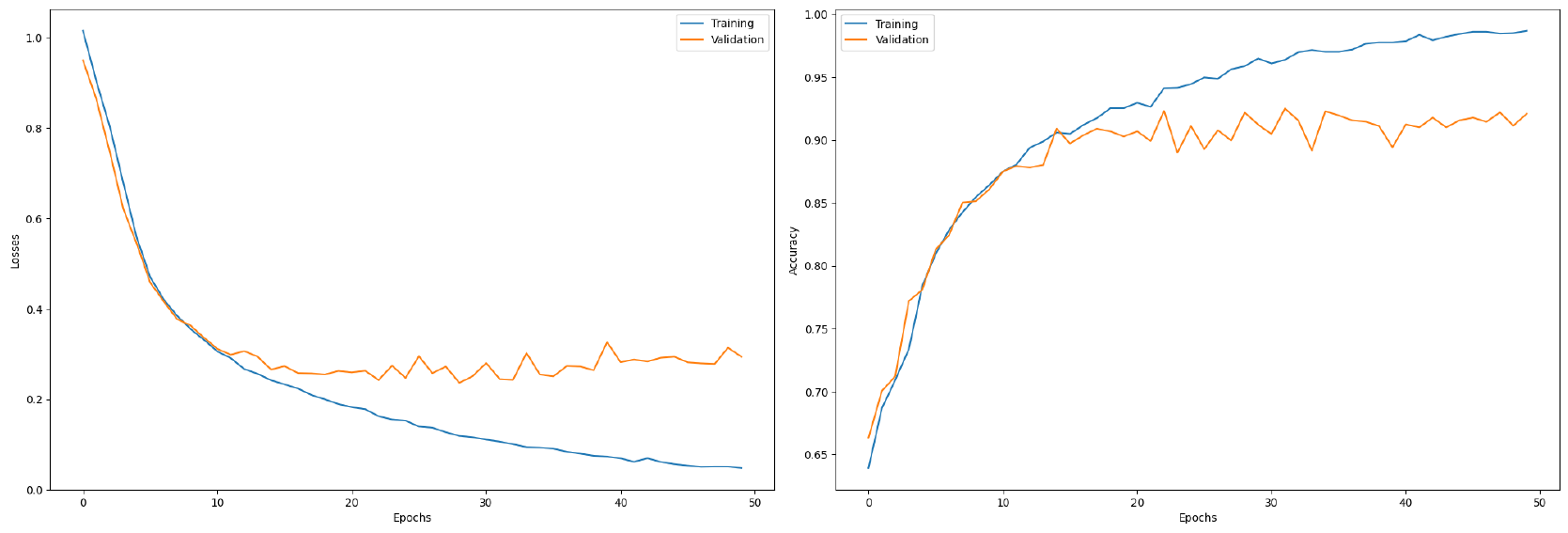}
    \caption{GAF-ViT Model Training and Validation Loss and Accuracy Variation over Epochs}
    \label{fig:fig7}
\end{figure}
Figure 8 displays the confusion matrix based on the testing set.
\begin{figure}
    \centering
    \includegraphics[width=9cm]{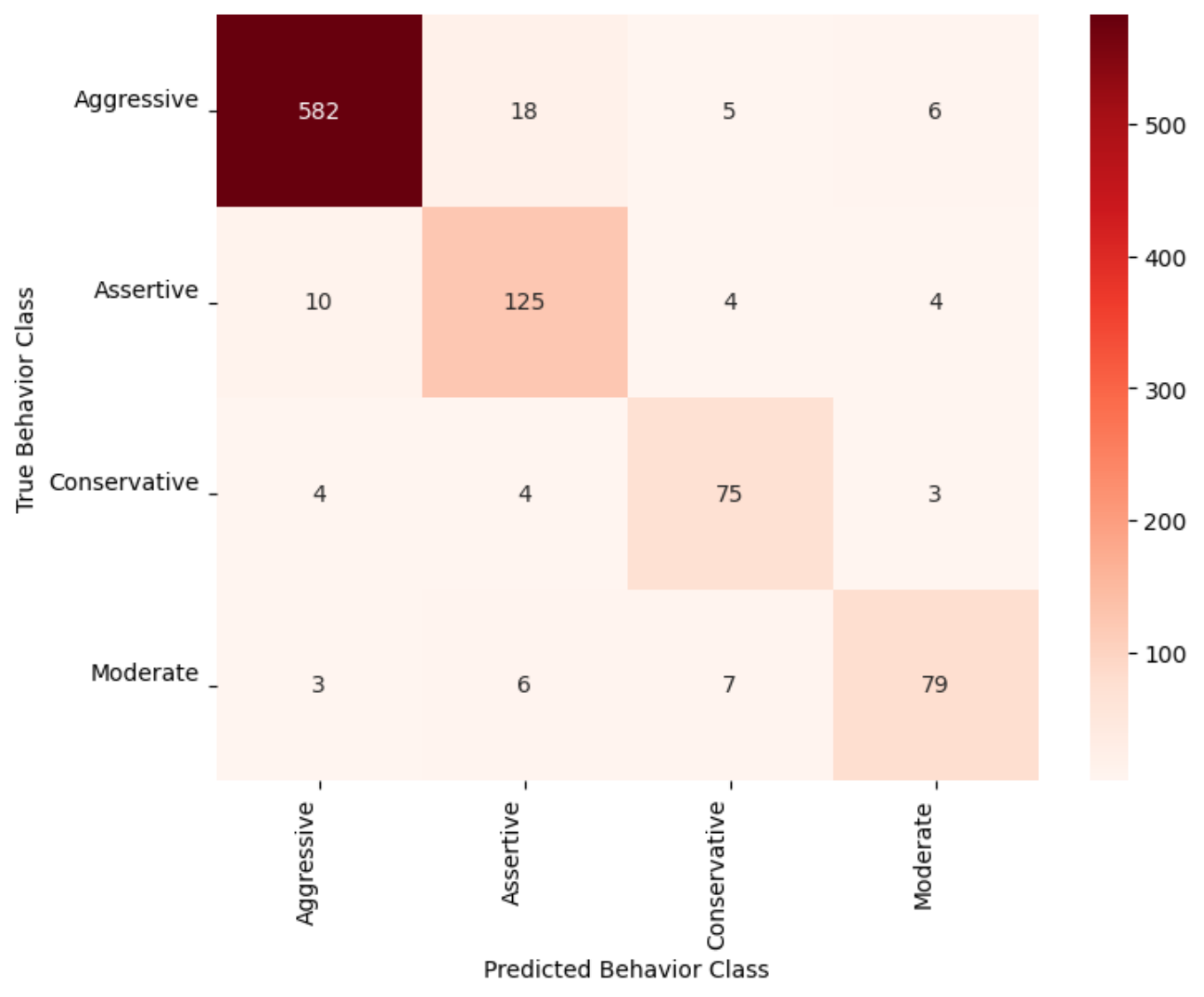}
    \caption{Confusion Matrix of GAF-ViT Model on Testing Set}
    \label{fig:fig8}
\end{figure}

\subsection{Model Comparison}
The developed GAF-ViT model is compared with the following benchmark models, widely adopted for multivariate time series classification:
\begin{itemize}
\item[$\bullet$] LSTM: Long Short-Term Memory \cite{graves2012long} is a type of recurrent neural network (RNN) designed to capture long-range dependencies within sequential data. Employing a memory cell alongside gates to regulate information flow, LSTMs prove efficacious for tasks involving sequential data, such as time series classification and prediction.
\end{itemize}
\begin{itemize}
\item[$\bullet$] MLP: Multilayer Perceptron \cite{wang2017time}, a feed-forward neural network characterized by multiple layers of interconnected neurons, stands out as a versatile model. It demonstrates capability across various tasks, including classification, regression, and function approximation.
\end{itemize}
\begin{itemize}
\item[$\bullet$] FCN: Fully Convolutional Network \cite{wang2017time} is primarily designed for image segmentation and related tasks. It replaces fully connected layers with convolutional layers, enabling it to accommodate input data of varied sizes.
\end{itemize}
\begin{itemize}
\item[$\bullet$] LSTM-FCN: LSTM-FCN Hybrid model \cite{karim2017lstm} combines the strengths of both LSTM and FCN architectures employing LSTM layers to discern temporal dependencies and utilizing FCN layers for adept feature extraction.
\end{itemize}
\begin{itemize}
\item[$\bullet$] GRU-FCN: Similar to LSTM-FCN, GRU-FCN Hybrid model \cite{elsayed2018deep} combines the GRU (Gated Recurrent Unit) with FCN layers for time series data classification.
\end{itemize}
\begin{itemize}
\item[$\bullet$] mWDN: Multiscale Weighted Dense Network \cite{wang2018multilevel} incorporates multiscale dilated convolutional layers and weighted dense connections to capture both local and global features in time series data for effective classification.
\end{itemize}
\begin{itemize}
\item[$\bullet$] MLSTM-FCN: Multiscale LSTM-FCN Hybrid model \cite{karim2019multivariate} combines LSTM and FCN layers with a multi-scale approach. It uses LSTM layers to capture temporal dependencies at different scales and FCN layers for feature extraction.
\end{itemize}
\begin{itemize}
\item[$\bullet$] TST: Time Series Transformer \cite{zerveas2021transformer} is based on the Transformer architecture and is designed specifically for time series data. It has demonstrated robust performance across various time series classification tasks by employing self-attention mechanisms to discern temporal dependencies.
\end{itemize}
\begin{itemize}
\item[$\bullet$] gMLP: Gated Multilayer Perceptron \cite{liu2021pay} represents a variation of the conventional MLP architecture, incorporating gated activation functions. This model introduces gating mechanisms into the MLP layers to enhance sequential data modeling.
\end{itemize}
The performance of the GAF-ViT model compared with the benchmark models in terms of Accuracy, Precision, Recall, and the F1 Score is shown in \uppercase\expandafter{\romannumeral3}.
\begin{table*}[!htbp]
\centering
\caption{Performance of GAF-ViT Model and Benchmark Models}\label{tab:aStrangeTable}
\setlength{\tabcolsep}{10mm}{
\begin{tabular}{ccccc}
\toprule
Model&Accuracy&Precision&Recall&F1\\
\bottomrule
LSTM     &0.7166&0.6227&0.4836&0.4955\\
MLP      &0.8321&0.8584&0.6721&0.7394\\
FCN      &0.8075&0.7915&0.6540&0.7040\\
LSTM-FCN &0.8032&0.8080&0.6334&0.6940\\
GRU-FCN  &0.6909&0.5536&0.4554&0.4782\\
mWDN     &0.9005&0.8595&0.8224&0.8385\\
MLSTM-FCN&0.8182&0.8003&0.6843&0.7311\\
TST      &0.7508&0.7896&0.4985&0.5586\\
gMLP     &0.8791&0.8344&0.8004&0.8164\\
\textbf{GAF-ViT}&\textbf{0.9209}&\textbf{0.8679}&\textbf{0.8826}&\textbf{0.8747}\\
\toprule
\end{tabular}}
\end{table*}
Table \uppercase\expandafter{\romannumeral3} presents experimental results, revealing that the proposed GAF-ViT model outperforms other baseline models across all evaluated metrics. Both wMDN and gMLP deliver relatively promising results through specially designed architectures to capture additional information from the time series. 

\subsection{Ablation Study}
An ablation study was conducted to further explore the impact on the performance associated with the removal of critical modules from the GAF-ViT model. Specifically, two ablated models were constructed to assess the indispensability of the Channel Attention Module and the GAF Transformation Module, respectively. Without the Channel Attention Module, multi-channel images resulting from the GAF Transformation Module are directly used for classification, implying that all driving behavior features are considered equal in defining a specific driving behavior type and classifying the resulting behavior. Conversely, without the GAF Transformation Module, the input 3D driving behavior matrix is reshaped to a size identical to the original multi-channel image through a linear layer for consistency before being fed through subsequent modules. The results from these two models are depicted in Figure 9. 
\begin{figure}
    \centering
    \includegraphics[width=9cm]{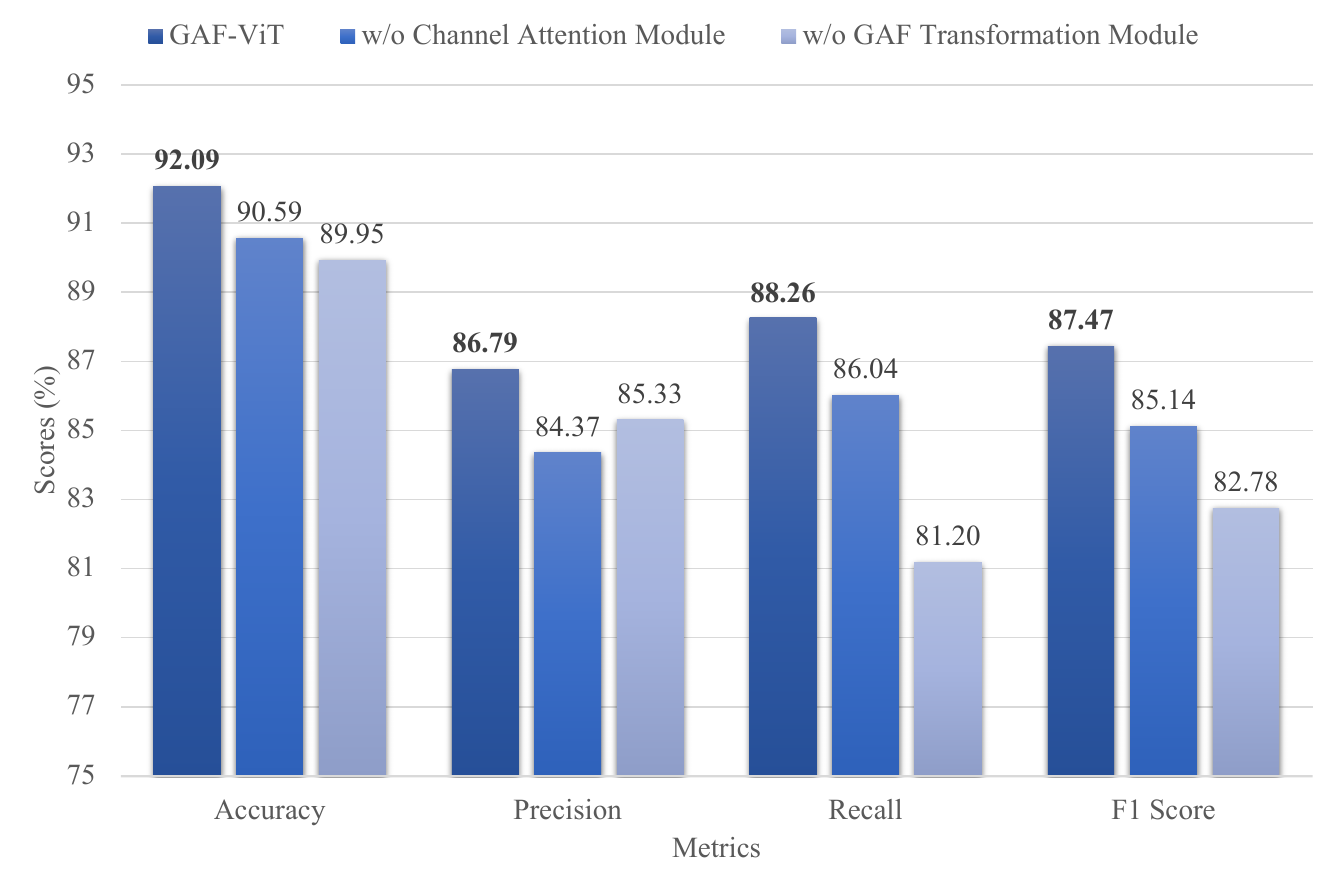}
    \caption{Results of Ablation Study}
    \label{fig:fig9}
\end{figure}
The results show that the performance of the original GAF-ViT model is better than both ablated models in terms of Accuracy, Precision, Recall, and the F1 Score, which confirm the necessity of both the Channel Attention Module and the GAF Transformation Module.
\section{Conclusion}
This study develops a Gramian Angular Field Vision Transformer (GAF-ViT) model for AV driving behavior imaging and classification. The model comprises three modules: GAF Transformation Module, Channel Attention Module, and Multi-Channel ViT Module. GAF Transformation Module converts multivariate driving behavior feature sequences into corresponding multi-channel images.  Subsequently. Channel Attention Module is designed to assign variant weights to different feature images across diverse channels, correlating to their importance. Ultimately, Multi-Channel ViT Module classifies AV driving behaviors by interpreting the weighted multi-channel images. Experimental outcomes indicate that the GAF-ViT Model outperforms benchmark models, widely used for classifying human driving behaviors. It demonstrates the immense potential for real-world deployment. This model stands to benefit research and validation teams by facilitating the detection of anomalies in AV operations and refining control algorithms.  This ensures the adaptive behavior of AVs in varied traffic scenarios, ultimately ensuring the safety of vehicles, passengers, and other participants while minimizing accident occurrences. 

Acknowledging the computational intensity of the GAF-ViT model due to the inherent demands of Vision Transformers, a practical deployment strategy involves leveraging edge cloud or roadside unit (RSU) infrastructure. These systems possess greater computational capabilities than the vehicle's onboard systems, enabling more flexible computational environments.  By offloading heavy data processing and inference tasks from the vehicle to these external units, the model can benefit from higher computational power and storage capacities, processing and analyzing driving behaviors more efficiently and with lower latency. This setup not only addresses computational constraints but also enhances the scalability of deploying advanced deep learning models like GAF-ViT in real-world autonomous driving applications. Such an arrangement facilitates continuous model improvement and updating without directly impacting the vehicle's operational efficiency, making it a viable solution for integrating sophisticated AI-driven behavior analysis into the autonomous driving ecosystem. Future research directions include: 1) Training and evaluating the model on more comprehensive datasets spanning a wider range of behavior features to enhance model generality and robustness, 2) Developing new modules that assimilate various data modalities, including those from cameras, LiDAR, or Radar-based visual and audio information, to further enhance the model's performance and efficacy.

\small
\bibliographystyle{IEEEtran}
\bibliography{IEEEabrv,refs}

\begin{thebibliography}{10}
\providecommand{\url}[1]{#1}
\csname url@samestyle\endcsname
\providecommand{\newblock}{\relax}
\providecommand{\bibinfo}[2]{#2}
\providecommand{\BIBentrySTDinterwordspacing}{\spaceskip=0pt\relax}
\providecommand{\BIBentryALTinterwordstretchfactor}{4}
\providecommand{\BIBentryALTinterwordspacing}{\spaceskip=\fontdimen2\font plus
\BIBentryALTinterwordstretchfactor\fontdimen3\font minus
  \fontdimen4\font\relax}
\providecommand{\BIBforeignlanguage}[2]{{%
\expandafter\ifx\csname l@#1\endcsname\relax
\typeout{** WARNING: IEEEtran.bst: No hyphenation pattern has been}%
\typeout{** loaded for the language `#1'. Using the pattern for}%
\typeout{** the default language instead.}%
\else
\language=\csname l@#1\endcsname
\fi
#2}}
\providecommand{\BIBdecl}{\relax}
\BIBdecl

\bibitem{muzahid2023multiple}
A.~J.~M. Muzahid, S.~F. Kamarulzaman, M.~A. Rahman, S.~A. Murad, M.~A.~S.
  Kamal, and A.~H. Alenezi, ``Multiple vehicle cooperation and collision
  avoidance in automated vehicles: survey and an ai-enabled conceptual
  framework,'' \emph{Scientific reports}, vol.~13, no.~1, p. 603, 2023.

\bibitem{wu2022human}
Z.~Wu, F.~Qu, L.~Yang, and J.~Gong, ``Human-like decision making for autonomous
  vehicles at the intersection using inverse reinforcement learning,''
  \emph{Sensors}, vol.~22, no.~12, p. 4500, 2022.

\bibitem{zhang2018human}
Y.~Zhang, P.~Sun, Y.~Yin, L.~Lin, and X.~Wang, ``Human-like autonomous vehicle
  speed control by deep reinforcement learning with double q-learning,'' in
  \emph{2018 IEEE intelligent vehicles symposium (IV)}.\hskip 1em plus 0.5em
  minus 0.4em\relax IEEE, 2018, pp. 1251--1256.

\bibitem{fu2019human}
R.~Fu, Z.~Li, Q.~Sun, and C.~Wang, ``Human-like car-following model for
  autonomous vehicles considering the cut-in behavior of other vehicles in
  mixed traffic,'' \emph{Accident Analysis \& Prevention}, vol. 132, p. 105260,
  2019.

\bibitem{lu2019personalized}
C.~Lu, J.~Gong, C.~Lv, X.~Chen, D.~Cao, and Y.~Chen, ``A personalized behavior
  learning system for human-like longitudinal speed control of autonomous
  vehicles,'' \emph{Sensors}, vol.~19, no.~17, p. 3672, 2019.

\bibitem{chen2022towards}
Z.~Chen, X.~Wang, Q.~Guo, and A.~Tarko, ``Towards human-like speed control in
  autonomous vehicles: A mountainous freeway case,'' \emph{Accident Analysis \&
  Prevention}, vol. 166, p. 106566, 2022.

\bibitem{huang2024human}
Z.~Huang, Z.~Sheng, C.~Ma, and S.~Chen, ``Human as ai mentor: Enhanced
  human-in-the-loop reinforcement learning for safe and efficient autonomous
  driving,'' \emph{Communications in Transportation Research}, vol.~4, p.
  100127, 2024.

\bibitem{mueller2020humanlike}
A.~S. Mueller, J.~B. Cicchino, and D.~S. Zuby, ``What humanlike errors do
  autonomous vehicles need to avoid to maximize safety?'' \emph{Journal of
  safety research}, vol.~75, pp. 310--318, 2020.

\bibitem{mit_news}
M.~C. Adam Conner-Simons, Rachel~Gordon, ``Predicting people's driving
  personalities,''
  \url{https://news.mit.edu/2019/predicting-driving-personalities-1118}.

\bibitem{hong2020driver}
Z.~Hong, Y.~Chen, and Y.~Wu, ``A driver behavior assessment and recommendation
  system for connected vehicles to produce safer driving environments through a
  “follow the leader” approach,'' \emph{Accident Analysis \& Prevention},
  vol. 139, p. 105460, 2020.

\bibitem{shahverdy2020driver}
M.~Shahverdy, M.~Fathy, R.~Berangi, and M.~Sabokrou, ``Driver behavior
  detection and classification using deep convolutional neural networks,''
  \emph{Expert Systems with Applications}, vol. 149, p. 113240, 2020.

\bibitem{melnyk2023driver}
Y.~Melnyk, S.~Otrokh, O.~Sarafannikov, and Y.~Lebid, ``Driver behavior
  recognition based on neural networks theory,'' \emph{Electronics and Control
  Systems}, vol.~1, no.~75, pp. 44--48, 2023.

\bibitem{zhang2022study}
R.~Zhang and X.~Ke, ``Study on distracted driving behavior based on transfer
  learning,'' in \emph{2022 IEEE 10th Joint International Information
  Technology and Artificial Intelligence Conference (ITAIC)}, vol.~10.\hskip
  1em plus 0.5em minus 0.4em\relax IEEE, 2022, pp. 1315--1319.

\bibitem{kuroki2021semi}
T.~Kuroki, O.~Shouno, and J.~Yoshimoto, ``Semi-supervised estimation of driving
  behaviors using robust time-contrastive learning,'' in \emph{2021
  Asia-Pacific Signal and Information Processing Association Annual Summit and
  Conference (APSIPA ASC)}.\hskip 1em plus 0.5em minus 0.4em\relax IEEE, 2021,
  pp. 1363--1366.

\bibitem{zhou2019analysis}
T.~Zhou and J.~Zhang, ``Analysis of commercial truck drivers’ potentially
  dangerous driving behaviors based on 11-month digital tachograph data and
  multilevel modeling approach,'' \emph{Accident Analysis \& Prevention}, vol.
  132, p. 105256, 2019.

\bibitem{hong2019rules}
J.~Hong, B.~Sapp, and J.~Philbin, ``Rules of the road: Predicting driving
  behavior with a convolutional model of semantic interactions,'' in
  \emph{Proceedings of the IEEE/CVF Conference on Computer Vision and Pattern
  Recognition}, 2019, pp. 8454--8462.

\bibitem{miglani2019deep}
A.~Miglani and N.~Kumar, ``Deep learning models for traffic flow prediction in
  autonomous vehicles: A review, solutions, and challenges,'' \emph{Vehicular
  Communications}, vol.~20, p. 100184, 2019.

\bibitem{tang2022prediction}
X.~Tang, K.~Yang, H.~Wang, J.~Wu, Y.~Qin, W.~Yu, and D.~Cao,
  ``Prediction-uncertainty-aware decision-making for autonomous vehicles,''
  \emph{IEEE Transactions on Intelligent Vehicles}, vol.~7, no.~4, pp.
  849--862, 2022.

\bibitem{wang2021intelligent}
W.~Wang, T.~Qie, C.~Yang, W.~Liu, C.~Xiang, and K.~Huang, ``An intelligent
  lane-changing behavior prediction and decision-making strategy for an
  autonomous vehicle,'' \emph{IEEE transactions on industrial electronics},
  vol.~69, no.~3, pp. 2927--2937, 2021.

\bibitem{mandal2020motion}
S.~Mandal, S.~Biswas, V.~E. Balas, R.~N. Shaw, and A.~Ghosh, ``Motion
  prediction for autonomous vehicles from lyft dataset using deep learning,''
  in \emph{2020 IEEE 5th International Conference on Computing Communication
  and Automation (ICCCA)}.\hskip 1em plus 0.5em minus 0.4em\relax IEEE, 2020,
  pp. 768--773.

\bibitem{luo2020probabilistic}
C.~Luo, L.~Sun, D.~Dabiri, and A.~Yuille, ``Probabilistic multi-modal
  trajectory prediction with lane attention for autonomous vehicles,'' in
  \emph{2020 IEEE/RSJ International Conference on Intelligent Robots and
  Systems (IROS)}.\hskip 1em plus 0.5em minus 0.4em\relax IEEE, 2020, pp.
  2370--2376.

\bibitem{jeong2020surround}
Y.~Jeong, S.~Kim, and K.~Yi, ``Surround vehicle motion prediction using
  lstm-rnn for motion planning of autonomous vehicles at multi-lane turn
  intersections,'' \emph{IEEE Open Journal of Intelligent Transportation
  Systems}, vol.~1, pp. 2--14, 2020.

\bibitem{wang2015imaging}
Z.~Wang and T.~Oates, ``Imaging time-series to improve classification and
  imputation,'' \emph{arXiv preprint arXiv:1506.00327}, 2015.

\bibitem{dosovitskiy2020image}
A.~Dosovitskiy, L.~Beyer, A.~Kolesnikov, D.~Weissenborn, X.~Zhai,
  T.~Unterthiner, M.~Dehghani, M.~Minderer, G.~Heigold, S.~Gelly \emph{et~al.},
  ``An image is worth 16x16 words: Transformers for image recognition at
  scale,'' \emph{arXiv preprint arXiv:2010.11929}, 2020.

\bibitem{hu2018squeeze}
J.~Hu, L.~Shen, and G.~Sun, ``Squeeze-and-excitation networks,'' in
  \emph{Proceedings of the IEEE conference on computer vision and pattern
  recognition}, 2018, pp. 7132--7141.

\bibitem{xing2017identification}
Y.~Xing, C.~Lv, Z.~Zhang, H.~Wang, X.~Na, D.~Cao, E.~Velenis, and F.-Y. Wang,
  ``Identification and analysis of driver postures for in-vehicle driving
  activities and secondary tasks recognition,'' \emph{IEEE Transactions on
  Computational Social Systems}, vol.~5, no.~1, pp. 95--108, 2017.

\bibitem{zhang2020driver}
C.~Zhang, R.~Li, W.~Kim, D.~Yoon, and P.~Patras, ``Driver behavior recognition
  via interwoven deep convolutional neural nets with multi-stream inputs,''
  \emph{Ieee Access}, vol.~8, pp. 191\,138--191\,151, 2020.

\bibitem{liu2020leveraging}
S.~Liu, R.~Muresan, and A.~Al-Dweik, ``Leveraging deep learning for inattentive
  driving behavior with in-vehicle cameras,'' in \emph{2020 International
  Symposium on Networks, Computers and Communications (ISNCC)}.\hskip 1em plus
  0.5em minus 0.4em\relax IEEE, 2020, pp. 1--6.

\bibitem{fan2022hybrid}
P.~Fan, J.~Guo, Y.~Wang, and J.~S. Wijnands, ``A hybrid deep learning approach
  for driver anomalous lane changing identification,'' \emph{Accident Analysis
  \& Prevention}, vol. 171, p. 106661, 2022.

\bibitem{fridman2019advanced}
L.~Fridman, D.~E. Brown, M.~Glazer, W.~Angell, S.~Dodd, B.~Jenik,
  J.~Terwilliger, A.~Patsekin, J.~Kindelsberger, L.~Ding \emph{et~al.}, ``Mit
  advanced vehicle technology study: Large-scale naturalistic driving study of
  driver behavior and interaction with automation,'' \emph{IEEE Access},
  vol.~7, pp. 102\,021--102\,038, 2019.

\bibitem{wang2017much}
W.~Wang, C.~Liu, and D.~Zhao, ``How much data are enough? a statistical
  approach with case study on longitudinal driving behavior,'' \emph{IEEE
  Transactions on Intelligent Vehicles}, vol.~2, no.~2, pp. 85--98, 2017.

\bibitem{wu2016novel}
M.~Wu, S.~Zhang, and Y.~Dong, ``A novel model-based driving behavior
  recognition system using motion sensors,'' \emph{Sensors}, vol.~16, no.~10,
  p. 1746, 2016.

\bibitem{lopez2012driver}
J.~O. L{\'o}pez, A.~C.~C. Pinilla \emph{et~al.}, ``Driver behavior
  classification model based on an intelligent driving diagnosis system,'' in
  \emph{2012 15th international IEEE conference on intelligent transportation
  systems}.\hskip 1em plus 0.5em minus 0.4em\relax IEEE, 2012, pp. 894--899.

\bibitem{zhang2016driver}
C.~Zhang, M.~Patel, S.~Buthpitiya, K.~Lyons, B.~Harrison, and G.~D. Abowd,
  ``Driver classification based on driving behaviors,'' in \emph{Proceedings of
  the 21st International Conference on Intelligent User Interfaces}, 2016, pp.
  80--84.

\bibitem{xie2019maneuver}
J.~Xie and M.~Zhu, ``Maneuver-based driving behavior classification based on
  random forest,'' \emph{IEEE Sensors Letters}, vol.~3, no.~11, pp. 1--4, 2019.

\bibitem{malik2023framework}
M.~Malik and R.~Nandal, ``A framework on driving behavior and pattern using
  on-board diagnostics (obd-ii) tool,'' \emph{Materials Today: Proceedings},
  vol.~80, pp. 3762--3768, 2023.

\bibitem{saleh2017driving}
K.~Saleh, M.~Hossny, and S.~Nahavandi, ``Driving behavior classification based
  on sensor data fusion using lstm recurrent neural networks,'' in \emph{2017
  IEEE 20th International Conference on Intelligent Transportation Systems
  (ITSC)}.\hskip 1em plus 0.5em minus 0.4em\relax IEEE, 2017, pp. 1--6.

\bibitem{moukafih2019aggressive}
Y.~Moukafih, H.~Hafidi, and M.~Ghogho, ``Aggressive driving detection using
  deep learning-based time series classification,'' in \emph{2019 IEEE
  international symposium on INnovations in intelligent SysTems and
  applications (INISTA)}.\hskip 1em plus 0.5em minus 0.4em\relax IEEE, 2019,
  pp. 1--5.

\bibitem{lee2023privacy}
C.-H. Lee and H.-C. Yang, ``A privacy-preserving learning method for analyzing
  hev driver’s driving behaviors,'' \emph{IEEE Access}, 2023.

\bibitem{gong2023sifdrivenet}
Y.~Gong, J.~Lu, W.~Liu, Z.~Li, X.~Jiang, X.~Gao, and X.~Wu, ``Sifdrivenet:
  Speed and image fusion for driving behavior classification network,''
  \emph{IEEE Transactions on Computational Social Systems}, 2023.

\bibitem{sharma2023kernelized}
O.~Sharma, N.~Sahoo, and N.~B. Puhan, ``Kernelized convolutional transformer
  network based driver behavior estimation for conflict resolution at
  unsignalized roundabout,'' \emph{ISA transactions}, vol. 133, pp. 13--28,
  2023.

\bibitem{vyas2022transdbc}
J.~Vyas, N.~Bhardwaj, D.~Das \emph{et~al.}, ``Transdbc: Transformer for
  multivariate time-series based driver behavior classification,'' in
  \emph{2022 International Joint Conference on Neural Networks (IJCNN)}.\hskip
  1em plus 0.5em minus 0.4em\relax IEEE, 2022, pp. 1--8.

\bibitem{qin2023cnn}
P.~Qin, H.~Li, Z.~Li, W.~Guan, and Y.~He, ``A cnn-lstm car-following model
  considering generalization ability,'' \emph{Sensors}, vol.~23, no.~2, p. 660,
  2023.

\bibitem{fan2019car}
P.~Fan, J.~Guo, H.~Zhao, J.~S. Wijnands, and Y.~Wang, ``Car-following modeling
  incorporating driving memory based on autoencoder and long short-term memory
  neural networks,'' \emph{Sustainability}, vol.~11, no.~23, p. 6755, 2019.

\bibitem{qin2023research}
P.~Qin, X.~Li, S.~Bin, F.~Wu, and Y.~Pang, ``Research on transformer and long
  short-term memory neural network car-following model considering data loss,''
  \emph{Mathematical Biosciences and Engineering}, vol.~20, no.~11, pp.
  19\,617--19\,635, 2023.

\bibitem{li2023automated}
Q.~Li, X.~Li, H.~Yao, Z.~Liang, and W.~Xie, ``Automated vehicle identification
  based on car-following data with machine learning,'' \emph{IEEE Transactions
  on Intelligent Transportation Systems}, 2023.

\bibitem{eren2012estimating}
H.~Eren, S.~Makinist, E.~Akin, and A.~Yilmaz, ``Estimating driving behavior by
  a smartphone,'' in \emph{2012 IEEE Intelligent Vehicles Symposium}.\hskip 1em
  plus 0.5em minus 0.4em\relax IEEE, 2012, pp. 234--239.

\bibitem{carlos2019smartphone}
M.~R. Carlos, L.~C. Gonz{\'a}lez, J.~Wahlstr{\"o}m, G.~Ram{\'\i}rez,
  F.~Mart{\'\i}nez, and G.~Runger, ``How smartphone accelerometers reveal
  aggressive driving behavior?—the key is the representation,'' \emph{IEEE
  Transactions on Intelligent Transportation Systems}, vol.~21, no.~8, pp.
  3377--3387, 2019.

\bibitem{lindow2019driver}
F.~Lindow and A.~Kashevnik, ``Driver behavior monitoring based on smartphone
  sensor data and machine learning methods,'' in \emph{2019 25th Conference of
  Open Innovations Association (FRUCT)}.\hskip 1em plus 0.5em minus 0.4em\relax
  IEEE, 2019, pp. 196--203.

\bibitem{khodairy2021driving}
M.~A. Khodairy and G.~Abosamra, ``Driving behavior classification based on
  oversampled signals of smartphone embedded sensors using an optimized
  stacked-lstm neural networks,'' \emph{IEEE Access}, vol.~9, pp. 4957--4972,
  2021.

\bibitem{brahim2022machine}
S.~B. Brahim, H.~Ghazzai, H.~Besbes, and Y.~Massoud, ``A machine learning
  smartphone-based sensing for driver behavior classification,'' in \emph{2022
  IEEE International Symposium on Circuits and Systems (ISCAS)}.\hskip 1em plus
  0.5em minus 0.4em\relax IEEE, 2022, pp. 610--614.

\bibitem{mantouka2021smartphone}
E.~Mantouka, E.~Barmpounakis, E.~Vlahogianni, and J.~Golias, ``Smartphone
  sensing for understanding driving behavior: Current practice and
  challenges,'' \emph{International journal of transportation science and
  technology}, vol.~10, no.~3, pp. 266--282, 2021.

\bibitem{xu2020human}
H.~Xu, J.~Li, H.~Yuan, Q.~Liu, S.~Fan, T.~Li, and X.~Sun, ``Human activity
  recognition based on gramian angular field and deep convolutional neural
  network,'' \emph{IEEE Access}, vol.~8, pp. 199\,393--199\,405, 2020.

\bibitem{hong2023monitoring}
S.~Hong, J.~Yoon, Y.~Ham, B.~Lee, and H.~Kim, ``Monitoring safety behaviors of
  scaffolding workers using gramian angular field convolution neural network
  based on imu sensing data,'' \emph{Automation in Construction}, vol. 148, p.
  104748, 2023.

\bibitem{garibo2023gramian}
{\`O}.~Garibo-i Orts, N.~Firbas, L.~Sebasti{\'a}, and J.~A. Conejero, ``Gramian
  angular fields for leveraging pretrained computer vision models with
  anomalous diffusion trajectories,'' \emph{Physical Review E}, vol. 107,
  no.~3, p. 034138, 2023.

\bibitem{jin2022delving}
X.~Jin, Y.~Xie, X.-S. Wei, B.-R. Zhao, Z.-M. Chen, and X.~Tan, ``Delving deep
  into spatial pooling for squeeze-and-excitation networks,'' \emph{Pattern
  Recognition}, vol. 121, p. 108159, 2022.

\bibitem{vaswani2017attention}
A.~Vaswani, N.~Shazeer, N.~Parmar, J.~Uszkoreit, L.~Jones, A.~N. Gomez,
  {\L}.~Kaiser, and I.~Polosukhin, ``Attention is all you need,''
  \emph{Advances in neural information processing systems}, vol.~30, 2017.

\bibitem{han2021transformer}
K.~Han, A.~Xiao, E.~Wu, J.~Guo, C.~Xu, and Y.~Wang, ``Transformer in
  transformer,'' \emph{Advances in Neural Information Processing Systems},
  vol.~34, pp. 15\,908--15\,919, 2021.

\bibitem{rao2021msa}
R.~M. Rao, J.~Liu, R.~Verkuil, J.~Meier, J.~Canny, P.~Abbeel, T.~Sercu, and
  A.~Rives, ``Msa transformer,'' in \emph{International Conference on Machine
  Learning}.\hskip 1em plus 0.5em minus 0.4em\relax PMLR, 2021, pp. 8844--8856.

\bibitem{kitaev2020reformer}
N.~Kitaev, {\L}.~Kaiser, and A.~Levskaya, ``Reformer: The efficient
  transformer,'' \emph{arXiv preprint arXiv:2001.04451}, 2020.

\bibitem{hu2022processing}
X.~Hu, Z.~Zheng, D.~Chen, X.~Zhang, and J.~Sun, ``Processing, assessing, and
  enhancing the waymo autonomous vehicle open dataset for driving behavior
  research,'' \emph{Transportation Research Part C: Emerging Technologies},
  vol. 134, p. 103490, 2022.

\bibitem{garyfallidis2012quickbundles}
E.~Garyfallidis, M.~Brett, M.~M. Correia, G.~B. Williams, and I.~Nimmo-Smith,
  ``Quickbundles, a method for tractography simplification,'' \emph{Frontiers
  in neuroscience}, vol.~6, p. 175, 2012.

\bibitem{blazquez2021clustering}
A.~Blazquez-Herranz, J.-I. Caballero-Garzon, A.~Zilverberg, C.~Wolff,
  A.~Rodr{\'\i}guez-Gonzalez, and E.~Menasalvas, ``Clustering moving object
  trajectories: Integration in cross-cpp analytic toolbox,'' \emph{Applied
  Sciences}, vol.~11, no.~8, p. 3693, 2021.

\bibitem{kumyaito2020trajectory}
N.~Kumyaito and K.~Tamee, ``Trajectory clustering by gps tracking dataset using
  quickbundles,'' \emph{ICIC express letters. Part B, Applications: an
  international journal of research and surveys}, vol.~11, no.~10, pp.
  921--928, 2020.

\bibitem{loshchilov2019decoupled}
I.~Loshchilov and F.~Hutter, ``Decoupled weight decay regularization,''
  \emph{arXiv preprint arXiv:1711.05101}, 2017.

\bibitem{graves2012long}
A.~Graves and A.~Graves, ``Long short-term memory,'' \emph{Supervised sequence
  labelling with recurrent neural networks}, pp. 37--45, 2012.

\bibitem{wang2017time}
Z.~Wang, W.~Yan, and T.~Oates, ``Time series classification from scratch with
  deep neural networks: A strong baseline,'' in \emph{2017 International joint
  conference on neural networks (IJCNN)}.\hskip 1em plus 0.5em minus
  0.4em\relax IEEE, 2017, pp. 1578--1585.

\bibitem{karim2017lstm}
F.~Karim, S.~Majumdar, H.~Darabi, and S.~Chen, ``Lstm fully convolutional
  networks for time series classification,'' \emph{IEEE access}, vol.~6, pp.
  1662--1669, 2017.

\bibitem{elsayed2018deep}
N.~Elsayed, A.~S. Maida, and M.~Bayoumi, ``Deep gated recurrent and
  convolutional network hybrid model for univariate time series
  classification,'' \emph{arXiv preprint arXiv:1812.07683}, 2018.

\bibitem{wang2018multilevel}
J.~Wang, Z.~Wang, J.~Li, and J.~Wu, ``Multilevel wavelet decomposition network
  for interpretable time series analysis,'' in \emph{Proceedings of the 24th
  ACM SIGKDD International Conference on Knowledge Discovery \& Data Mining},
  2018, pp. 2437--2446.

\bibitem{karim2019multivariate}
F.~Karim, S.~Majumdar, H.~Darabi, and S.~Harford, ``Multivariate lstm-fcns for
  time series classification,'' \emph{Neural networks}, vol. 116, pp. 237--245,
  2019.

\bibitem{zerveas2021transformer}
G.~Zerveas, S.~Jayaraman, D.~Patel, A.~Bhamidipaty, and C.~Eickhoff, ``A
  transformer-based framework for multivariate time series representation
  learning,'' in \emph{Proceedings of the 27th ACM SIGKDD conference on
  knowledge discovery \& data mining}, 2021, pp. 2114--2124.

\bibitem{liu2021pay}
H.~Liu, Z.~Dai, D.~So, and Q.~V. Le, ``Pay attention to mlps,'' \emph{Advances
  in Neural Information Processing Systems}, vol.~34, pp. 9204--9215, 2021.

\end{thebibliography}

\begin{IEEEbiography}[{\includegraphics[width=1in,height=1.25in,clip,keepaspectratio]{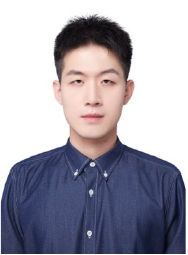}}]{Junwei You}
received the M.S. degree in Civil and Environmental Engineering from Northwestern University in 2022. He is currently a Ph.D. student in Civil and Environmental Engineering at University of Wisconsin-Madison. His research interests include multimodal generative AI, connected and automated vehicles, and deep learning advancement in intelligent transportation systems.
\end{IEEEbiography}

\begin{IEEEbiography}[{\includegraphics[width=1in,height=1.25in,clip,keepaspectratio]{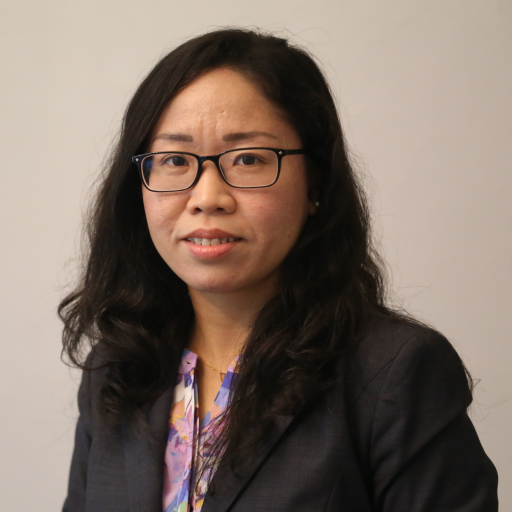}}]{Ying Chen}
is an Assistant Professor of Instruction in the Department of Civil and Environmental Engineering at Northwestern University. She specializes in solving problems by using AI, machine learning, and data mining techniques, with an emphasis on data science in transportation and topics related to smart cities, complex networks analysis, modeling, and solution approaches for logistics and complex systems, connected/autonomous vehicles, social networking influence, and big social media data mining for travelers’ behavior study. Dr.Chen received her Ph.D. in Civil and Environmental Engineering and M.S. in Computer Science from Northwestern University in 2015 and 2011.
\end{IEEEbiography}

\begin{IEEEbiography}
[{\includegraphics[width=1in,height=1.25in,clip,keepaspectratio]{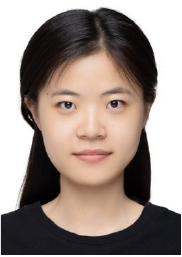}}]{Zhuoyu Jiang}
received her B.S. degrees in Mathematics and Statistics from University of Wisconsin-Madison in 2023. Her research interests include foundation models, generative AI, and data mining. 
\end{IEEEbiography}

\begin{IEEEbiography}
[{\includegraphics[width=1in,height=1.25in,clip,keepaspectratio]{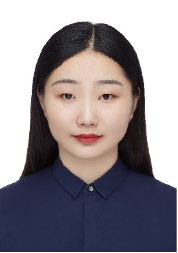}}]{Zhangchi Liu}
received the B.S. degree in Transportation Engineering from Beijing Jiaotong University, China, in 2020, and the M.S. degree in Civil and Environmental Engineering from Northwestern University in 2022. Her research interests include data analytics in transportation and travel behavior analysis.
\end{IEEEbiography}

\begin{IEEEbiography}[{\includegraphics[width=1in,height=1.25in,clip,keepaspectratio]{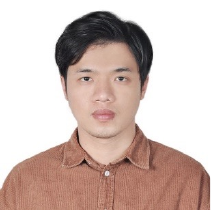}}]{Zilin Huang}
received his B.S. degree from the School of Electromechanical Engineering, Guangdong University of Technology in 2018. He received his M.S. degree in Communication and Transportation Engineering from South China University of Technology in 2021. He is currently pursuing a Ph.D. degree at the Department of Civil and Environmental Engineering, University of Wisconsin-Madison, USA. Before joining UW-Madison, he worked at the Center for Connected and Automated Transportation (CCAT), Purdue University, USA. His research interests include human-centered AI, autonomous driving, robotics, human-robot interaction, and intelligent transportation.
\end{IEEEbiography}

\begin{IEEEbiography}
[{\includegraphics[width=1in,height=1.25in,clip,keepaspectratio]{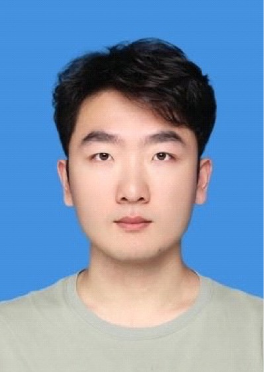}}]{Yifeng Ding}
received the B.S. degree in Software Engineering from Nankai University, China, in 2020, and the M.S. degree in Computer Techonology from Tsinghua University, China, in 2023. His research interests include Natural Language Processing and Multimodal Pre-Training.
\end{IEEEbiography}

\begin{IEEEbiography}
[{\includegraphics[width=1in,height=1.25in,clip,keepaspectratio]{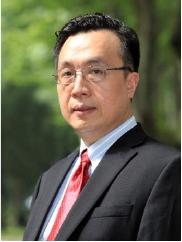}}]{Bin Ran}
is the Vilas Distinguished Achievement Professor and Director of ITS Program at University of Wisconsin-Madison. Dr. Ran is an expert in dynamic transportation network models, traffic simulation and control, traffic information system, internet of mobility, and Connected Automated Vehicle Highway (CAVH) systems. He has led the development and deployment of various traffic information systems and the demonstration of CAVH systems. Dr. Ran is the author of two leading textbooks on dynamic traffic networks. He has co-authored more than 240 journal papers and more than 260 referenced papers at national and international conferences. He holds more than 20 patents of CAVH in the US and other countries. He is an associate editor of Journal of Intelligent Transportation Systems.
\end{IEEEbiography}

\end{document}